\documentclass[11pt,a4paper]{article}
\usepackage[hyperref]{emnlp2020}

\usepackage{microtype}
\usepackage{times}
\usepackage{url}
\usepackage{epsfig}
\usepackage{graphicx}
\usepackage{amsmath}
\usepackage{amssymb}
\usepackage{color}
\usepackage{kky}
\usepackage{todonotes}
\usepackage{multirow}
\usepackage{latexsym}
\usepackage{hyperref}
\usepackage{xcolor}
\usepackage{booktabs}
\usepackage{soul}
\usepackage{adjustbox}
\usepackage{makecell}

\usepackage{microtype}

\aclfinalcopy 
\def\aclpaperid{***} 

\newcommand{\model}{CAST}

\newcommand{\xv}{\mathbf{x}}
\newcommand{\yv}{\mathbf{y}}
\newcommand{\cv}{\mathbf{c}}
\newcommand{\pv}{\mathbf{p}}
\newcommand{\uv}{\mathbf{u}}

\newcommand{\E}{\mathbb{E}}

\title{Contextual Text Style Transfer}

\author{Yu Cheng$^1$, Zhe Gan$^1$, Yizhe Zhang$^1$, Oussama Elachqar$^1$, Dianqi Li$^2$, Jingjing Liu$^1$
\vspace{1mm} \\
\hspace{0.15in} $^{1}$Microsoft
\hspace{0.15in} $^{2}$University of Washington
\vspace{1mm} \\
\small\tt{\{yu.cheng,zhe.gan,yizhe.zhang,ouelachq,jinjl\}@microsoft.com},
\small\tt{dianqili@uw.edu}
}

\date{}

\begin{document}
\maketitle
\begin{abstract}
We introduce a new task, Contextual Text Style Transfer - translating a sentence into a desired style with its surrounding context taken into account. This brings two key challenges to existing style transfer approaches: ($i$) how to preserve the semantic meaning of target sentence and its consistency with surrounding context during transfer; ($ii$) how to train a robust model with limited labeled data accompanied with context. To realize high-quality style transfer with natural context preservation, we propose a Context-Aware Style Transfer (CAST) model, which uses two separate encoders for each input sentence and its surrounding context. A classifier is further trained to ensure contextual consistency of the generated sentence. To compensate for the lack of parallel data, additional self-reconstruction and back-translation losses are introduced to leverage non-parallel data in a semi-supervised fashion. Two new benchmarks, Enron-Context and Reddit-Context, are introduced for formality and offensiveness style transfer. Experimental results on these datasets demonstrate the effectiveness of the proposed CAST model over state-of-the-art methods across style accuracy, content preservation and contextual consistency metrics.
\end{abstract}

\section{Introduction}
Text style transfer has been applied to many applications (e.g., sentiment manipulation, formalized writing) with remarkable success. Early work relies on parallel corpora
with a sequence-to-sequence learning framework \citep{BahdanauCB14,jhamtani2017shakespearizing}. However, collecting parallel annotations is highly time-consuming and expensive. There has also been studies on developing text style transfer models with non-parallel data \citep{hu2017toward,li2018delete,prabhumoye2018style,subramanian2018multiple}, assuming that disentangling style information from semantic content can be achieved in an auto-encoding fashion with the introduction of additional regularizers (e.g., adversarial discriminators \citep{shen2017style}, language models \citep{yang2018unsupervised}). 

Despite promising results, these techniques still have a long way to go for practical use. Most existing models focus on sentence-level rewriting. However, in real-world applications, sentences typically reside in a surrounding paragraph context.
In formalized writing, the rewritten span is expected to align well with the surrounding context to keep a coherent semantic flow. For example, to automatically replace a gender-biased sentence in a job description document, a style transfer model taking the sentence out of context may not be able to understand the proper meaning of the statement and the intended message. Taking a single sentence as the sole input of a style transfer model may fail in preserving topical coherency between the generated sentence and its surrounding context, leading to low semantic and logical consistency on the paragraph level (see Example C in Table~\ref{tab:example}). Similar observations can be found in other style transfer tasks, such as offensive to non-offensive and political to neutral translations. 

Motivated by this, we propose and investigate a new task - \emph{Contextual Text Style Transfer}. Given a paragraph, the system aims to translate sentences into a desired style, while keeping the edited section topically coherent with its surrounding context. To achieve this goal, we propose a novel Context-Aware Style Transfer (CAST) model, by jointly considering style translation and context alignment. To leverage parallel training data, CAST employs two separate encoders to encode the source sentence and its surrounding context, respectively. With the encoded sentence and context embeddings, a decoder is trained to translate the joint features into a new sentence in a specific style. A pre-trained style classifier is applied for style regularization, and a coherence classifier learns to regularize the generated target sentence to be consistent with the context. To overcome data sparsity issue, we further introduce a set of unsupervised training objectives (e.g., self-reconstruction loss, back-translation loss) to leverage non-parallel data in a hybrid approach. 
The final CAST model is jointly trained with both parallel and non-parallel data via end-to-end training.

As this is a newly proposed task, we introduce
two new datasets, \emph{Enron-Context} and \emph{Reddit-Context}, collected via crowdsourcing. The former contains 14,734 formal vs. informal paired samples from Enron~\citep{Klimt2004IntroducingTE} (an email dataset), and the latter contains 23,158 offensive vs. non-offensive paired samples from Reddit~\citep{DBLP:journals/corr/abs-1709-02349}. Each sample contains an original sentence and a human-rewritten one in target style, accompanied by its paragraph context. In experiments, we also leverage 60k formal/informal sentences from GYAFC~\citep{rao2018dear} and 100k offensive/non-offensive sentences from Reddit~\citep{dosSantos:acl2018} as additional non-parallel data for model training. 

The main contributions of this work are summarized as follows:
($i$) We propose a new task - Contextual Text Style Transfer, which aims to translate a sentence into a desired style while preserving its style-agnostic semantics and topical consistency with the surrounding context.
($ii$) We introduce two new datasets for this task, Enron-Context and Reddit-Context, which provide strong benchmarks for evaluating contextual style transfer models. 
($iii$) We present a new model - Context-Aware Style Transfer (CAST), which jointly optimizes the generation quality of target sentence and its topical coherency with adjacent context. Extensive experiments on the new datasets demonstrate that the proposed CAST model significantly outperforms state-of-the-art style transfer models.

\section{Related Work}
\subsection{Text Style Transfer} 

Text style transfer aims to modify an input sentence into a desired style while preserving its style-independent semantics. Previous work has explored this as a sequence-to-sequence learning task using parallel corpora with paired source/target sentences in different styles. For example,
\citet{jhamtani2017shakespearizing} pre-trained word embeddings by leveraging external dictionaries mapping Shakespearean words to modern English words and additional text. However, available parallel data in different styles are very limited. Therefore, there is a recent surge of interest in considering a more realistic setting, where only non-parallel stylized corpora are available. A typical approach is: ($i$) disentangling latent space as content and style features; then ($ii$) generating stylistic sentences by tweaking style-relevant features and passing them through a decoder, together with the original content-relevant features~\citep{xu2018unpaired}. 

Many of these approaches borrowed the idea of adversarial discriminator/classifier from the Generative Adversarial
Network (GAN) framework \citep{gan}. For example, 
\citet{shen2017style,fu2018style,lample2017unsupervised} used adversarial classifiers to force the decoder to transfer the encoded source sentence into a different style/language. Alternatively, \citet{li2018delete} achieved disentanglement by filtering stylistic words of input sentences. Another direction for text style transfer without parallel data is using back-translation~\citep{prabhumoye2018style} with a de-noising auto-encoding objective~\citep{logeswaran2018content,subramanian2018multiple}.

\begin{figure*}[t!]
\centering
\includegraphics[width=0.99\linewidth]{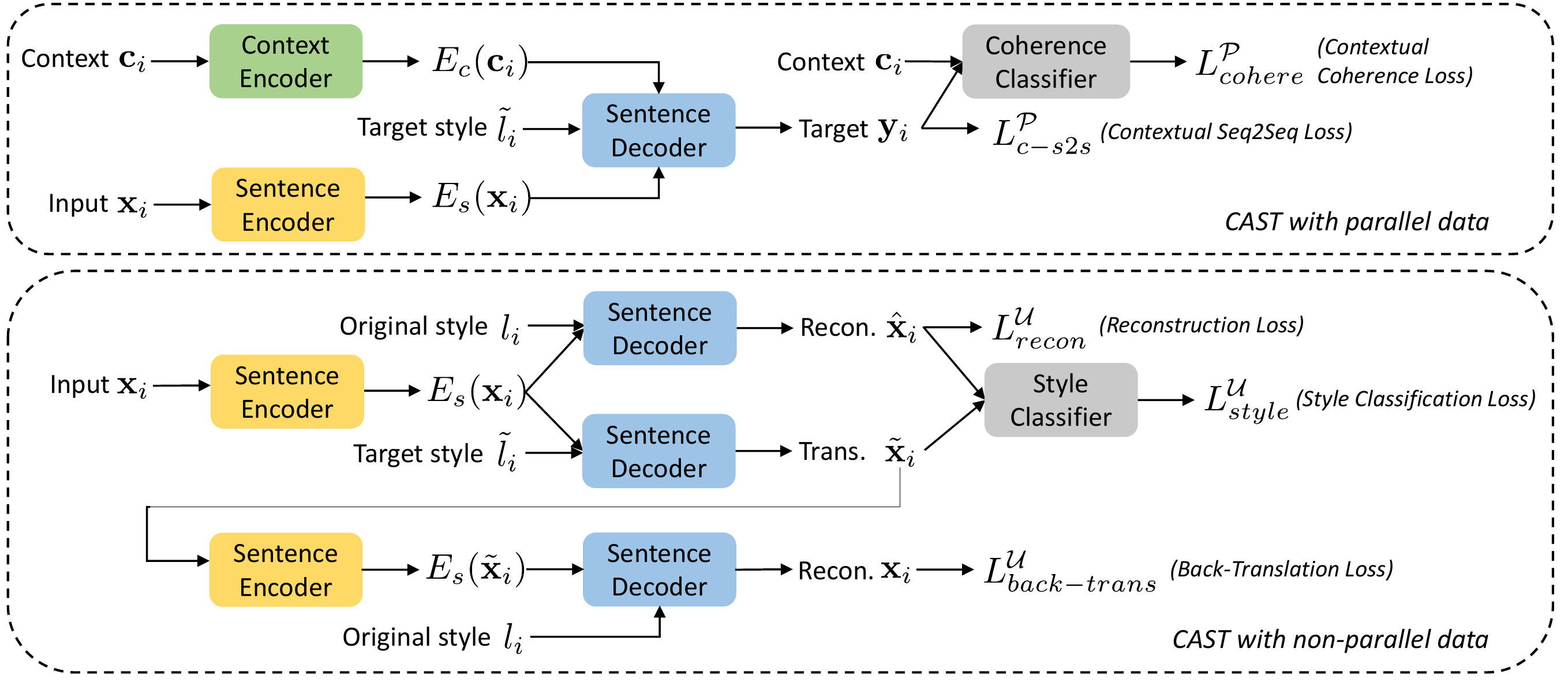}
\caption{Model architecture of the proposed CAST model for contextual text style transfer. Both the training paths share the same sentence encoder and decoder. See Sec.~\ref{ref:cast} for details.}
\label{fig:model_arch}
\end{figure*}

Regarding the tasks, sentiment transfer is one of the most widely studied problems. From informality to formality \citep{rao2018dear,li2019domian} is another direction of text style transfer, aiming to change the style of a given sentence to more formal text. \citet{dosSantos:acl2018} presented an approach to transferring offensive text to non-offensive based on social network data. In \citet{prabhumoye2018style}, the authors proposed the political slant transfer task. However, all these previous studies did not directly consider context-aware text style transfer, which is the main focus of this work.

\subsection{Context-aware Text Generation} 

Our work is related to context-aware text generation \citep{Mikolov2012ContextDR,Tang2016ContextawareNL}, which can be applied to many NLP tasks \citep{mangrulkar-etal-2018-context}. For example, previous work has investigated language modeling with context information~\citep{DBLP:journals/corr/WangC15g,wang2017topic}, treating the preceding sentences as context.
There are also studies on response generation for conversational systems \citep{sordoni-etal-2015-neural,wen-etal-2015-stochastic}, where dialogue history is treated as a context. \citet{zang-wan-2017-towards} introduced a neural model to generate long reviews from aspect-sentiment scores given the topics. \citet{DBLP:journals/corr/VinyalsL15} proposed a model to predict the next sentence given the previous sentences in a dialogue session. \citet{Sordoni:2015:HRE:2806416.2806493} presented a hierarchical recurrent encoder-decoder model to encode dialogue context. Our work is the first to explore context information in the text style transfer task.

\section{Context-Aware Style Transfer} \label{ref:cast}


In this section, we first describe the problem definition and provide an overview of the model architecture in Section~\ref{sec:problem_definition}. Section~\ref{sec:cast_with_parallel} presents the proposed Context-Aware Style Transfer (CAST) model with supervised training objectives, and Section~\ref{sec:cast_with_nonparallel} further introduces how to augment the CAST model with non-parallel data in a hybrid approach.

\subsection{Overview}

\paragraph{Problem Definition} \label{sec:problem_definition}
The problem of contextual text style transfer is defined as follows.
A style-labelled parallel dataset $\Pcal = \{(\xv_i,l_i), (\yv_i,\tilde{l}_i), \cv_i\}^{M}_{i = 1}$ includes: ($i$) the $i$-th instance containing the original sentence $\mathbf{x}_i$ with a style $l_i$, ($ii$) its corresponding rewritten sentence $\mathbf{y}_i$ in another style $\tilde{l}_i$, and ($iii$) the paragraph context $\mathbf{c}_i$. $\mathbf{x}_i$ and $\mathbf{y}_i$ are expected to encode the same semantic content, but in different language styles (i.e., $l_i\not= \tilde{l}_i$). The goal is to transform $\mathbf{x}_i$ in style $l_i$ to $\mathbf{y}_i$ in style $\tilde{l}_i$, while keeping $\mathbf{y}_i$ semantically coherent with its context $\mathbf{c}_i$. In practice, labelled parallel data may be difficult to garner. Ideally, additional non-parallel data $\mathcal{U} = \{(\mathbf{x}_i, l_i)\}^{N}_{i=1}$ can be leveraged to enhance model training.

\paragraph{Model Architecture}
The architecture of the proposed CAST model is illustrated in Figure \ref{fig:model_arch}. The hybrid model training process consists of two paths, one for parallel data and the other for non-parallel data. In the parallel path, a \emph{Seq2Seq loss} and a \emph{contextual coherence loss} are included, for the joint training of two encoders (Sentence Encoder and Context Encoder) and the Sentence Decoder. The non-parallel path is designed to further enhance the Sentence Encoder and Decoder with three additional losses: ($i$) a \emph{self-reconstruction loss}; ($ii$) a \emph{back-translation loss}; and ($iii$) a \emph{style classification loss}.  
The final training objective, uniting both parallel and non-parallel paths, is formulated as:
\begin{equation}
\begin{split}
    L_{final}^{\mathcal{P}, \mathcal{U}} = L_{c-s2s}^{\mathcal{P}} + \lambda_1 L_{cohere}^{\mathcal{P}} + \lambda_2  L_{recon}^{\mathcal{U}} \\ + \lambda_3 L_{btrans}^{\mathcal{U}} + \lambda_4 L_{style}^{\mathcal{U}}\,,
    \label{eq:all}    
\end{split}
\end{equation}
where $\lambda_1$, $\lambda_2$, $\lambda_3$ and $\lambda_4$ are hyper-parameters to balance different objectives. Each of these loss terms will be explained in the following sub-sections.

\subsection{Supervised Training Objectives} \label{sec:cast_with_parallel}
In this subsection, we discuss the training objective associated with parallel data, consisting of: ($i$) a contextual Seq2Seq loss; and ($ii$) a contextual coherence loss.

\paragraph{Contextual Seq2Seq Loss}
When parallel data are available, a Seq2Seq model can be directly learned for text style transfer. We denote the Seq2Seq model as $(E,D)$, where the semantic representation of sentence $\xv_i$ is extracted by the encoder $E$, and the decoder $D$ aims to learn a conditional distribution of $\yv_i$ given the encoded feature $E(\xv_i)$ and style $\tilde{l}_i$:
\begin{equation}
    L_{s2s}^{\mathcal{P}} = -\mathop{\E}_{\xv_i,\yv_i \sim \mathcal{P}} \log p_D(\yv_i|E(\mathbf{x}_i), \tilde{l}_i)\,.
    \label{eq:seq2seq}
\end{equation}
However, in such a sentence-to-sentence style transfer setting, the context in the paragraph is ignored, which if well utilized, could help improve generation quality such as paragraph-level topical coherence.

Thus, to take advantage of the paragraph context $\mathbf{c}_i$, we use two separate encoders $E_s$ and $E_c$ to encode the sentence and the context independently. The outputs of the two encoders are combined via a linear layer, to obtain a context-aware sentence representation, which is then fed to the decoder to generate the target sentence. The model is trained to minimize the following loss:
\begin{equation}
    L_{c-s2s}^{\mathcal{P}} = -\mathop{\mathbb{E}}_{\mathbf{x}_i, \cv_i, \yv_i  \sim \mathcal{P}} \log p_D (\mathbf{y}_i | E_s(\mathbf{x}_i), E_c(\mathbf{c}_i), \tilde{l}_i)\,.
    \label{eq:context_seq2seq}
\end{equation}
Compared with Eqn.~(\ref{eq:seq2seq}), the use of $E_c(\mathbf{c}_i)$ makes the text style transfer process context-dependent. The generated sentence can be denoted as $\tilde{\yv}_i = D(E_s(\xv_i),E_c(\cv_i),\tilde{l}_i)$.

\paragraph{Contextual Coherence Loss}
To enforce contextual coherence (i.e., to ensure the generated sentence $\yv_i$ aligns with the surrounding context $\cv_i$), we train a coherence classifier that judges whether $\mathbf{c}_i$ is the context of $\mathbf{y}_i$, by adopting a language model with an objective similar to next sentence prediction~\citep{bert}. 

Specifically, assume that $\yv_i$ is the $t$-th sentence of a paragraph $\pv_i$ (i.e., $\yv_i=\pv_i^{(t)}$), and $\cv_i = \{\pv_i^{(0)},\ldots,\pv_i^{(t-1)},\pv_i^{(t+1)},\ldots,\pv_i^{(T)}\}$ is its surrounding context. We first reconstruct the paragraph $\pv_i = \{\pv_i^{(0)},\ldots,\pv_i^{(T)}\}$ by inserting $\yv_i$ into the proper position in $\cv_i$, denoted as $[\cv_i;\yv_i]$. Based on this, we obtain a paragraph representation $\uv_i$ via a language model encoder. Then, we apply a linear layer to the representation, followed by a $\tanh$ function and a softmax layer to predict a binary label $s_i$, which indicates whether $\cv_i$ is the right context for $\yv_i$ :
\begin{align}
    \uv_{i} & = \text{LM}([\mathbf{c}_i; f(\mathbf{y}_{i})]) \\
    p_\text{LM}(s_i|\mathbf{c}_i, \mathbf{y}_{i}) & = \text{softmax} \rbr{\tanh \rbr{\Wb \ub_{i} + \bb}} \,, \nonumber
\end{align}
where \text{LM} represents the language model encoder, and $s_i=1$ indicates that $\mathbf{c}_i$ is the context of $\mathbf{y}_i$. Note that since $\tilde{\yv}_i$ are discrete tokens that are non-differentiable, we use the continuous feature $f(\tilde{\yv}_i)$ to generates $\tilde{\yv}_i$ as the input of the language model.
We construct paired data $\{\yv_i,\cv_i,s_i\}_{i=1}^N$ for training the classifier, where the negative samples are created by replacing a sentence in a paragraph with another random sentence. After pre-training, the coherence classifier is used to obtain the contextual coherence loss:
\begin{align}
    L_{cohere}^{\mathcal{P}}  = -\mathop{\mathbb{E}}_{\mathbf{x}_i, \cv_i \sim \mathcal{P}} \log p_\text{LM}(s_i=1|\mathbf{c}_i,f(\tilde{\yv}_i)) \,.
    \label{eq:coherence}
\end{align}
%
Intuitively, minimizing $L_{cohere}^{\mathcal{P}}$ encourages $\tilde{\yv}_i$ to blend better to its context $\mathbf{c}_i$. Note that the coherence classifier is pre-trained, and remains fixed during the training of the CAST model. The above coherence loss can be used to update the parameters of $E_s, E_c$ and $D$ during model training. 


\subsection{Unsupervised Training Objectives} \label{sec:cast_with_nonparallel}
For the contextual style transfer task, there are not many parallel datasets available with style-labeled paragraph pairs. 
To overcome the data sparsity issue, we propose a hybrid approach to leverage additional non-parallel data $\mathcal{U}=\{(\mathbf{x}_i, l_i)\}^{N}_{i=1}$, which are abundant and less expensive to collect. 
In order to fully exploit $\mathcal{U}$ to enhance the training of the Sentence Encoder and Decoder $(E_s,D)$, we introduce three additional training losses, detailed below.


\paragraph{Reconstruction Loss} The reconstruction loss aims to encourage $E_s$ and $D$ to reconstruct the input sentence itself, if the desired style is the same as the input style. The corresponding objective is similar to Eqn.~(\ref{eq:seq2seq}):
\begin{equation}
    L_{recon}^{\mathcal{U}} = -\mathop{\mathbb{E}}_{\mathbf{x}_i \sim \mathcal{U}} \log p_D (\mathbf{x}_i | E_s(\mathbf{x}_i), l_i)\,.
    \label{eq:recon}
\end{equation}
Compared to Eqn.~(\ref{eq:seq2seq}), here we encourage the decoder $D$ to recover $\xv_i$'s original style properties as accurate as possible, given the style label $l_i$. The self-reconstructed sentence is denoted as $\hat{\mathbf{x}}_i = D(E_s(\xv_i),l_i)$.


\paragraph{Back-Translation Loss}
The back-translation loss requires the
model to reconstruct the input sentence after a
transformation loop. Specifically, the input sentence $\xv_i$ is first transferred into the target style, i.e., $\tilde{\xv}_i = D(E_s(\xv_i),\tilde{l}_i)$. Then the generated target sentence is transferred back into its original style, i.e., $\hat{\xv}_i=D(E_s(\tilde{\xv}_i),l_i)$. 
The back-translation loss is defined as:
\begin{equation}
    L_{btrans}^{\mathcal{U}} = -\mathop{\mathbb{E}}_{\substack{\mathbf{x}_i \sim \mathcal{U},\tilde{\xv}_i \sim \\  p_D(\yv_i|E_s(\xv_i),\tilde{l}_i)}}  \log p_D (\mathbf{x}_i | E_s(\tilde{\xv}_i), l_i)\,,
    \label{eq:style}
\end{equation}
where the source and target styles are denoted as $l_i$ and $\tilde{l}_i$, respectively.

\paragraph{Style Classification Loss}
To further boost the model, we use $\mathcal{U}$ to train a classifier that predicts the style of a given sentence, and regularize the training of $(E_s,D)$ with the pre-trained style classifier. The objective is defined as:
\begin{equation}
    L_{style} = -\mathop{\mathbb{E}}_{\mathbf{x}_i \sim \mathcal{U}} \log p_{C}(l_i | \mathbf{x}_i)\,,
\end{equation}
where $p_{C}(\cdot)$ denotes the style classifier. After the classifier is trained, we keep its parameters
fixed, and apply it to update the parameters of $(E_s,D)$. 
The resulting style classification loss utilizing the pre-trained style classifier is defined as:
\begin{equation}
\begin{split}
    L_{style}^\mathcal{U} = -\mathop{\mathbb{E}}_{\mathbf{x}_i \sim \mathcal{U}}\Big[ \mathop{\mathbb{E}}_{\hat{\xv}_i \sim p_D(\hat{\xv}_i|E_s(\xv_i),l_i)} \log p_{C}(l_i | \hat{\xv}_i) \\
    +\mathop{\mathbb{E}}_{\tilde{\xv}_i \sim p_D(\tilde{\xv}_i|E_s(\xv_i),\tilde{l}_i)}
    \log p_{C}(\tilde{l}_i | \tilde{\xv}_i) \Big] \,.
\end{split}
\end{equation}


\section{New Benchmarks} \label{sec:datasets}
Existing text style transfer datasets, either parallel or non-parallel, do not contain contextual information, thus unsuitable for the contextual transfer task. To provide benchmarks for evaluation, we introduce two new datasets: Enron-Context and Reddit-Context, derived from two existing datasets - Enron~\citep{Klimt2004IntroducingTE} and Reddit Politics~\citep{DBLP:journals/corr/abs-1709-02349}.  


\begin{table*}
\begin{center}
\small
\begin{tabular}{c|c|c|c|c|c}
\toprule
Dataset & \# sent.  & \# rewritten sent. & \# words per sent. & \# words per paragraph & \# vocabulary \\ 
\midrule 
Reddit-Context  & 14,734 & 14,734 & 9.4 & 38.5 & 4,622 \\ \midrule
Enron-Context &  23,158 & 25,259  & 7.6  & 25.9 & 2,196 \\ 
\bottomrule
\end{tabular}
\end{center}
\vspace{-1mm}
\caption{Statistics on Enron-Context and Reddit-Context datasets.}\label{table:stat:dataset}
\end{table*}

\paragraph{$1$) Enron-Context}
To build a formality transfer dataset with paragraph contexts, we randomly sampled emails from the Enron corpus \citep{Klimt2004IntroducingTE}. After pre-processing and filtering with NLTK \citep{BirdKleinLoper09}, we asked Amazon Mechanical Turk (AMT) annotators to identify informal sentences within each email, and rewrite them in a more formal style. Then, we asked a different group of annotators to verify if each rewritten sentence is more formal than the original sentence.

\paragraph{$2$) Reddit-Context} Another typical style transfer task is offensive vs. non-offensive, for which we collected another dataset from the Reddit Politics corpus \citep{DBLP:journals/corr/abs-1709-02349}. First, we identify offensive sentences in the original dataset with sentence-level classification. After filtering out extremely long/short sentences, we randomly selected a subset of sentences (10\% of the whole dataset) and asked AMT annotators to rewrite each offensive sentence into two non-offensive alternatives. 

After manually removing wrong or duplicate annotations, we obtained a total of 14,734 rewritten sentences for Enron-Context, and 23,158 for Reddit-Context. We also limited the vocabulary size by replacing words with a frequency less than 20/70 in Enron/Reddit datasets with a special unknown token. 
Table \ref{table:stat:dataset} provides the statistics on the two datasets. More details on AMT data collection are provided in Appendix.

\begin{table*}[t!]
\centering
\small{
\begin{tabular}{c|c|c||c|c|c|c|c}
\toprule 
\multicolumn{8}{c}{Formality Transfer}\\\midrule 
  Non-parallel   & Train & Style classifier & Parallel    & Train   & Dev & Test & Coherence classifier \\\midrule
  \textsc{GYAFC} & 58k & 12k & Enron-Context & 13k & 0.5k & 1k & 2.5k \\\midrule \midrule
  \multicolumn{8}{c}{Offensiveness Transfer}\\\midrule 
  Non-parallel  & Train  & Style classifier & Parallel  & Train   & Dev & Test & Coherence classifier\\\midrule
  \textsc{Reddit} & 106k & 15k & Reddit-Context & 22k & 0.5k & 1k & 3.5k\\\bottomrule
\end{tabular}
}
\vspace{-1mm}
\caption{Statistics of the parallel and non-parallel datasets on the two text style transfer tasks.}
\label{tab:data}
\end{table*}


\section{Experiments}
In this section, we compare our model with state-of-the-art baselines on the two new benchmarks, and provide both quantitative analysis and human evaluation
to validate the effectiveness of the proposed CAST model. 

%
\subsection{Datasets and Baselines}

In addition to the two new parallel datasets, we also leverage non-parallel datasets for CAST model training. For formality transfer, one choice is Grammarly’s Yahoo Answers Formality Corpus (GYAFC) \citep{rao2018dear}, crawled and annotated from two domains in Yahoo Answers. This corpus contains paired informal-formal sentences without context. We randomly selected a subset of sentences (28,375/29,774 formal/informal) from the GYAFC dataset as our training dataset. For offensiveness transfer, we utilize the Reddit dataset. Following \citet{dosSantos:acl2018}, we used a pre-trained classifier to extract 53,028/53,714 offensive/non-offensive sentences from Reddit posts as our training dataset. 

Table \ref{tab:data} provides the statistics of parallel and non-parallel datasets used for the two style transfer tasks. For the non-parallel datasets, we split them into two: one for CAST model training (`Train'), and the other for the style classifier pre-training. Similarly, for the parallel datasets, the training sets are divided into two as well, for the training of \model~ (`Train/Dev/Test') and the coherence classifier, respectively. 

We compare CAST model with several baselines: ($i$) Seq2Seq: a Transformer-based Seq2Seq model (Eqn.~(\ref{eq:seq2seq})), taking sentences as the only input, trained on parallel data only; ($ii$) Contextual Seq2Seq: a Transformer-based contextual Seq2Seq model (Eqn.~(\ref{eq:context_seq2seq})), taking both context and sentence as input, trained on parallel data only; ($iii$) Hybrid Seq2Seq~\citep{DBLP:journals/corr/abs-1903-06353}: a Seq2Seq model leveraging both parallel and non-parallel data; 
($iv$) ControlGen~\citep{hu2017toward,hu2018texar}: a state-of-the-art text transfer model using non-parallel data; ($v$) MulAttGen~\citep{subramanian2018multiple}: another state-of-the-art style transfer model that allows flexible control over multiple attributes. 

\subsection{Evaluation Metrics} The contextual style transfer task requires a model to generate sentences that: ($i$) preserve the original semantic content and structure in the source sentence; ($ii$) conform to the pre-specified style; and ($iii$) align with the surrounding context in the paragraph. Thus, we consider the following automatic metrics for evaluation: 

\paragraph{Content Preservation.} We assess the degree of content preservation during transfer, by measuring \textit{BLEU} scores \citep{papineni2002bleu} between generated sentences and human references. Following \citet{rao2018dear}, we also use \textit{GLEU} as an additional metric for the formality transfer task, which was originally introduced for the grammatical error correction task \citep{napoles-etal-2015-ground}. For offensiveness transfer, we include perplexity (\textit{PPL}) as used in \citet{dosSantos:acl2018}, which is computed by a word-level LSTM language model pre-trained on non-offensive sentences. 

\paragraph{Style Accuracy.} Similar to prior work, we measure style accuracy using the prediction accuracy of the pre-trained style classifier over generated sentences (\textit{Acc.}). 

\paragraph{Context Coherence.} We use the prediction accuracy of the pre-trained coherence classifier to measure how a generated sentence matches its surrounding context.


For formality transfer, the style classifier and coherence classifier reach 91.35\% and 86.78\% accuracy, respectively, on pre-trained dataset. 
For offensiveness transfer, the accuracy is 93.47\% and 84.96\%. Thus, we consider these measurements as reliable evaluation metrics for this task.

\subsection{Implementation Details} 
The context encoder, sentence encoder and sentence decoder are all implemented as a one-layer Transformer with 4 heads. The hidden dimension of one head is 256, and the hidden dimension of the feed-forward sub-layer is 1024. The context encoder is set to take maximum of 50 words from the surrounding context of the target sentence. For the style classifier, we use a standard CNN-based sentence classifier~\citep{kim2014convolutional}.

\begin{table*}[t!]
\centering
\small{
\begin{tabular}{ccccc|cccc}
\toprule
{} & \multicolumn{4}{c}{Formality Transfer} & \multicolumn{4}{c}{Offensiveness Transfer}\\\midrule
Model  &  \textit{Acc.} & \textit{Coherence} & \textit{BLEU} & \textit{GLEU}  & \textit{Acc.} & \textit{Coherence} & \textit{BLEU} & \textit{PPL} \\\midrule 
Seq2Seq & 64.05 & 78.09 & 24.16 & 10.46 & 83.05 & 80.28 & 17.22 & 140.39 \\
Contextual Seq2Seq & 64.28 & 81.25 & 23.72 & 10.37 & 83.42 & 81.69 & 18.74 & 138.42 \\
Hybrid Seq2Seq  & 65.09 & 79.62 & 24.35 & 10.93 & 83.28 & 84.87 & 20.78 & 107.12 \\
ControlGen & 62.18 & 73.66 & 14.32 & 8.72 & 82.15 & 78.81 & 10.44 & 92.14 \\
MulAttGen & 63.36 & 72.97 & 15.14 & 8.91 & 82.71 & 78.45 & 11.03 & \textbf{92.56} \\
\model  &  \textbf{68.04} & \textbf{85.47} & \textbf{26.38} & \textbf{15.06} & \textbf{88.45} & \textbf{85.98} & \textbf{23.92} & 93.03 \\
\bottomrule
\end{tabular}
}
\vspace{-1mm}
\caption{Quantitative evaluation results of different models on the two style transfer tasks.}
\label{tab:res}
\end{table*}

\begin{table*}[t!]
\centering
\small
\begin{adjustbox}{scale=0.94,tabular=ll|c|p{1.4in},center}
\toprule
{} & {} & \textbf{Task: informal to formal transfer}  & \textbf{Context} \\\midrule
\multirow{5}{*}{A} 
& \textbf{Input}
&\textcolor{orange}{I'm assuming that you'd set up be part of that meeting ?} &  \\
& \textbf{ControlGen}
&I'm guessing that you would be set up that call ? &
{I'll call him back to a } \\
& \textbf{MulAttGen}
&I'm guessing that you would be set up that meeting ? &
{meeting.} {\color{orange}[Input].}  {I asked} \\
& \textbf{C-Seq2Seq}
&I am assuming that you would part of that person .& {him what sort of deals}\\
& \textbf{H-Seq2Seq}
&I am assuming that you would be part of that party ?& {they're working on .}\\ 
& \textbf{\model}
&Am I correct to assume that you would attend that \textcolor{blue}{meeting} ? & {}\\ \midrule 
\multirow{5}{*}{B} &
\textbf{Input}
&\textcolor{orange}{Do y'all interface with C/P .}  & {Thanks . Can someone let }
\\
& \textbf{ControlGen}
&Do you compete with them ? & {the C/P know that the deals} 
\\
& \textbf{MulAttGen}
&Do you interface with them ? &
{are good ?} {\color{orange}[Input].} {If not }\\
& \textbf{C-Seq2Seq}
&Do we interface with them ? & {deal confirmations could but}
\\
& \textbf{H-Seq2Seq}
&Do we interface with them ?  & {they need the deal details .}
\\
& \textbf{\model}
&Do you all interface with \textcolor{blue}{C/P} ?  & \\\midrule  \midrule
{} & {} &  \textbf{Task: offensive to non-offensive transfer}  & \textbf{Context} \\\midrule 
\multirow{5}{*}{C} &
\textbf{Input}
&\textcolor{orange}{You are ugly .} & {} 
\\
& \textbf{ControlGen}
&You bad guy ! & {With the glasses ,} {\color{orange}[Input].} \\
& \textbf{MulAttGen}
&You are sad . &
{I don't need them because I }\\
& \textbf{C-Seq2Seq}
&Have a bad day . & {never read . How do i look ?}
\\
& \textbf{H-Seq2Seq}
&What a bad day !  &  
\\
& \textbf{\model}
&You \textcolor{blue}{look} not good . & \\\bottomrule
\end{adjustbox}
\vspace{-1mm}
\caption{Examples from the two datasets, where {\color{orange}orange} denotes the sentence to be transferred, and {\color{blue}blue} denotes content that also appears in the context. \textbf{C-Seq2Seq}: Contextual Seq2Seq; \textbf{H-Seq2Seq}: Hybrid Seq2Seq.}
\label{tab:example}
\end{table*}

Since the non-parallel corpus $\mathcal{U}$ contains more samples than the parallel one $\mathcal{P}$, we down-sample $\mathcal{U}$ to assign each mini-batch the same number of parallel and non-parallel samples to balance training, alleviating the 'catastrophic forgetting problem' described in \citet{howard-ruder-2018-universal}.
We train the model using Adam optimizer with a mini-batch size 64 and a learning rate 0.0005. The validation set is used to select the best hyper-parameters. 
Hard-sampling \citep{logeswaran2018content} is used to back-propagate loss through discrete tokens from the pre-trained classifier to the model. 

For the ControlGen \citep{hu2017toward} baseline, we use the code provided by the authors, and use their default hyper-parameter setting. For Hybrid Seq2Seq~\citep{DBLP:journals/corr/abs-1903-06353} and MulAttGen~\citep{subramanian2018multiple}, we re-implement their models following the original papers. 

\subsection{Experimental Results}

\paragraph{Formality Transfer}
Results on the formality transfer task are summarized in Table \ref{tab:res}. The CAST model achieves better performance than all the baselines. Particularly, CAST is able to boost \textit{GLEU} and \textit{Coherence} scores with a large margin. Hybrid Seq2Seq also achieves good performance by utilizing non-parallel data. By incorporating context information, Contextual Seq2Seq also improves over the vanilla Seq2Seq model. 
As expected, ControlGen does not perform well, since only non-parallel data is used for training. 

\vspace{-2mm}
\paragraph{Offensiveness Transfer}
Results are summarized in Table \ref{tab:res}. CAST achieves the best performance over all the metrics except for \textit{PPL}. In terms of \textit{Coherence}, Contextual Seq2Seq and CAST, that leverage context information achieve better performance than Seq2Seq baseline. Contextual Seq2Seq also improves \textit{BLEU}, which is different from the observation in the formality transfer task. 
On \textit{PPL}, CAST produces slightly worse performance than ControlGen and MulAttGen. We hypothesize that this is because our model tends to use the same non-offensive word to replace an offensive word, producing some untypical sentences, as discussed in \citet{dosSantos:acl2018}. 

\begin{table*}[t!]
\centering
\small{
\begin{tabular}{ccccc|cccc}
\toprule
{} & \multicolumn{4}{c}{Formality Transfer} & \multicolumn{4}{c}{Offensiveness Transfer}\\\midrule
Model  &  \textit{Acc.} & \textit{Coherence} & \textit{BLEU} & \textit{GLEU}  & \textit{Acc.} & \textit{Coherence} & \textit{BLEU} & \textit{PPL} \\\midrule 
\model   &  \textbf{68.04} & \textbf{85.47} & \textbf{26.38} & \textbf{15.06} & \textbf{88.45} & \textbf{85.98} & \textbf{23.92} & \textbf{93.03} \\
w/o context encoder & 65.35 & 82.9 & 23.98 & 14.17 & 84.15 & 80.96 & 20.54 & 127.02\\
w/o cohere. classifier & 65.47 & 80.16 & 14.82 & 14.45 & 85.11 & 79.37 & 21.97 & 115.57\\
w/o both & 62.19 & 74.47 & 15.88 & 10.46 & 72.69 & 78.15 & 13.14  & 147.31 \\
w/o non-parallel data & 60.19 & 75.49 & 13.5 & 9.88 & 70.84 & 78.72 & 10.53 & 151.08 \\ \bottomrule
\end{tabular}
}
\caption{Ablation study of CAST on two style transfer tasks. }
\label{tab:abl}
\end{table*}

\begin{table*}
\centering
\small{
\begin{adjustbox}{scale=0.98,tabular=c|c|ccc|ccc|ccc,center}
\toprule
\multirow{2}{*}{Task} &
\multirow{2}{*}{Aspects} &
\multicolumn{3}{c|}{\model~vs.} & \multicolumn{3}{c|}{\model ~vs.} & \multicolumn{3}{c}{\model~vs.} \\ 
& & \multicolumn{3}{c|}{Contextual Seq2Seq} & \multicolumn{3}{c|}{Hybrid Seq2Seq} & \multicolumn{3}{c}{ControlGen} \\ \midrule

& & win & lose & tie & win & lose & tie & win & lose & tie \\
\midrule
\multirow{3}{*}{\makecell{Formality \\ Transfer}} & Style Control & \textbf{57.1} & 28.3 & 14.6 & \textbf{46.9} & 26.1 & 28.0 & \textbf{72.1} & 12.6 & 25.3\\
& Content Preservation & \textbf{59.7} & 22.1 & 18.2 & \textbf{50.4} & 20.8 & 28.2 & \textbf{68.8} & 14.5 & 17.7 \\
& Context Consistence & \textbf{56.4} & 23.1 & 20.5 & \textbf{51.5} & 19.7 & 28.8 & \textbf{70.1} & 10.6 & 19.3 \\
\midrule
\multirow{3}{*}{\makecell{ Offensiveness \\ Transfer}} & Style Control & \textbf{58.6} & 25.3 & 16.1 & \textbf{50.1} & 29.2 & 20.3 & \textbf{54.8} & 19.9 & 25.3\\
& Content Preservation & \textbf{62.3} & 26.5 & 11.2 & \textbf{54.0} & 17.5 & 28.5 & \textbf{53.1} & 30.2 & 16.7 \\
& Context Consistence & \textbf{60.1} & 32.4 & 17.5 & \textbf{55.3} & 24.9 & 20.8 & \textbf{58.1} & 35.8 & 16.7 \\
\bottomrule
\end{adjustbox}
}
\caption{Results of pairwise human evaluation between \model~and three baselines on two style transfer tasks. Win/lose/tie indicate the percentage of results generated by \model~being better/worse/equal to the reference model.}\label{tab:human_eval}
\end{table*}

\vspace{-2mm}
\paragraph{Qualitative Analysis}
Table \ref{tab:example} presents some generation examples from different models. 
We observe that CAST is better at replacing informal words with formal ones (Example B and C), and generates more context-aware sentences (Example A and C), possibly due to the use of coherence and style classifiers. We also observe that the exploitation of context information can help the model preserve semantic content in the original sentence (Example B). 

\vspace{-2mm}
\paragraph{Ablation Study}
To investigate the effectiveness of each component of CAST model, we conduct detailed ablation studies and summarize the results in Table \ref{tab:abl}. Experiments show that the context encoder and the coherence classifier play an important role in the proposed model. The context encoder is able to improve content preservation and style transfer accuracy, demonstrating the effectiveness of using context.
The coherence classifier can help improve the coherence score but not much for style accuracy. By using these two components, our model can strike a proper balance between translating to the correct style and maintaining contextual consistency.
When both of them are removed (the $4$th row), performance on all the metrics drops significantly. 
We also observe that without using non-parallel data, the model performs poorly, showing the benefit of using a hybrid approach and more data for this task. 

\vspace{-2mm}
\paragraph{Human Evaluation}
Considering the subjective nature of this task, we conduct human evaluation to judge model outputs regarding content preservation, style control and context consistency. 
Given an original sentence along with its corresponding context and a pair of generated sentences from two different models, AMT workers were asked to select the best one based on these three aspects. The AMT interface also allows a neutral option, if the worker considers both sentences as equally good in certain aspect. We randomly sampled 200 sentences from the test set, and collected three human responses for each pair. 
Table \ref{tab:human_eval} reports the pairwise comparison results on both tasks. Based on human judgment, the quality of transferred sentences by \model~is significantly higher than the other methods across all three metrics. This is consistent with the experimental results on automatic metrics discussed earlier. 

\section{Conclusion}
In this paper, we present a new task - Contextual Text Style Transfer. Two new benchmark datasets are introduced for this task,  which contain annotated sentence pairs accompanied by paragraph context. We also propose a new CAST model, which can effectively enforce content preservation and context coherence, by exploiting abundant non-parallel data in a hybrid approach. Quantitative and human evaluations demonstrate that CAST model significantly outperforms baseline methods that do not consider context information. 
We believe our model takes a first step towards modeling context information for text style transfer, and will explore more advanced solutions e.g., using a better encoder/decoder like GPT-2 \cite{radford2019language} and BERT \cite{bert}, adversarial learning \cite{Zhu2020FreeLB} or knowledge distillation \cite{distllbert}.

%
\bibliography{emnlp2020}
\bibliographystyle{acl_natbib}

\section{Appendix}
\paragraph{Data Collection:} We use the offensive language and hate speech classifier from \citet{davidson2017automated} to classify the offensive sentence. We perform the classification at the sentence level for each Reddit post. 

\begin{figure*}
\centering
\includegraphics[width=\linewidth]{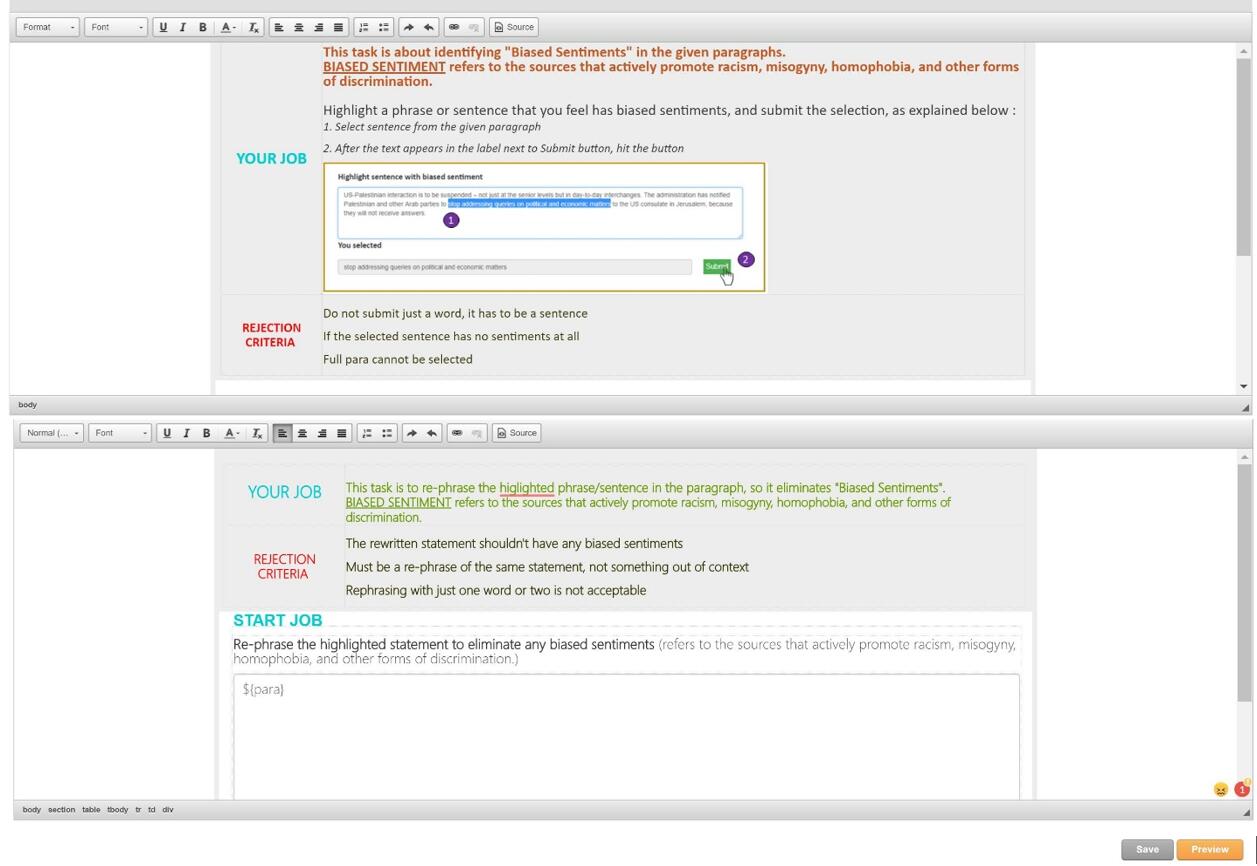}
\vspace{-3mm}
\caption{The AMT interfaces we used to collect the bias/violence data. The top interface is for turkers to select the bias/violence data from a paragraph. The bottom interface is for turkers to rewrite the sentence/phrase given the context.}
\label{fig:amt}
\vspace{-3mm}
\end{figure*}

As shown in Figure \ref{fig:amt}, the data collection are divided into two sub-tasks: first select the bias/violence data from a paragraph and then rewrite the sentence/phrase given the context.

The rewritten sentences from Enron-Context are validated by one of collaborators from a company. The 23,158 Reddit-Context are validated ourselves. Each rewritten sentence is reviewed by one volunteer to check if it is inoffensive while preserves the original content.

\paragraph{AMT Interface:} In Figure \ref{fig:amt}, we show the AMT user interfaces to collect the bias/violence data.

\end{document}